\documentclass[10pt,twocolumn,letterpaper]{article}

\usepackage{iccv}              

\newcommand{\modelName}{Prompt-DINO\xspace}
\newcommand{\rapLongName}{Recognize Anything via Prompting\xspace}
\newcommand{\rapName}{RAP\xspace}

\definecolor{iccvblue}{rgb}{0.21,0.49,0.74}
\usepackage[pagebackref,breaklinks,colorlinks,allcolors=iccvblue]{hyperref}
\usepackage{multirow}
\usepackage{float}

\def\paperID{10391} 
\def\confName{ICCV}
\def\confYear{2025}

\title{Text-guided Visual Prompt DINO for Generic Segmentation}

\author{Yuchen Guan\textsuperscript{1$\ast$$\ddagger$},
Chong Sun\textsuperscript{2$\ast$$\dagger$},
Canmiao Fu\textsuperscript{2$\ast$},
Zhipeng Huang\textsuperscript{2}, 
Chun Yuan\textsuperscript{1$\dagger$}, 
Chen Li\textsuperscript{2}\\
\textsuperscript{1}Tsinghua Shenzhen International Graduate School, Tsinghua University\\
\textsuperscript{2}WeChat AI, Tencent Inc.\\
}
\vspace{-5em}
\begin{document}

\twocolumn[{%
\renewcommand\twocolumn[1][]{#1}%
\maketitle

\begin{center}
    \vspace{-2em} 
    \includegraphics[width=1.0\linewidth]{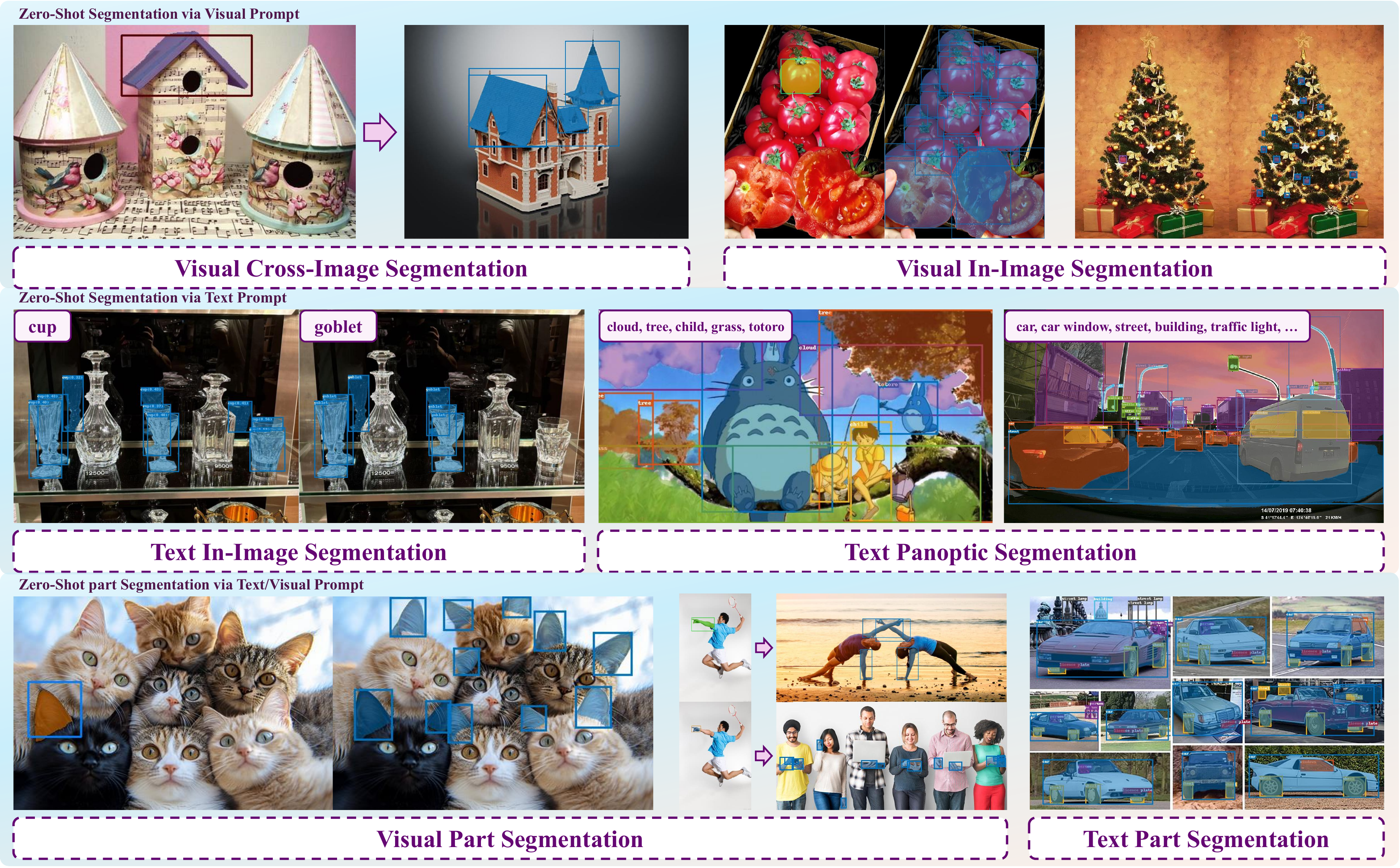}
    \vspace{-2em} 
    \captionof{figure}{
    \textbf{Detection/Segmentation samples.} 
    Our \modelName is a text-guided visual Prompt DINO. \modelName supports single or multiple visual or text prompts for in-image or cross-image generic segmentation. Based on the granularity of the provided prompts, \modelName is capable of performing both panoptic and part segmentation.
    }
    \label{fig:teaser}
\end{center}%
}]

\renewcommand{\thefootnote}{}
\footnotetext{$^\ast$ Equal Contribution.}
\footnotetext{$^\dagger$ Corresponding Authors. Emails: waynecsun@tencent.com, \\ 
\hangindent=2.5em yuanc@sz.tsinghua.edu.cn}
\footnotetext{$^\ddagger$ Work done as interns at WeChat.}

\vspace{-2em} 
\begin{abstract}
Recent advancements in multimodal vision models have highlighted limitations in late-stage feature fusion and suboptimal query selection for hybrid prompts open-world segmentation, alongside constraints from caption-derived vocabularies. To address these challenges, we propose \modelName, a text-guided visual Prompt DINO framework featuring three key innovations. First, we introduce an early fusion mechanism that unifies text/visual prompts and backbone features at the initial encoding stage, enabling deeper cross-modal interactions to resolve semantic ambiguities. Second, we design order-aligned query selection for DETR-based architectures, explicitly optimizing the structural alignment between text and visual queries during decoding to enhance semantic-spatial consistency. Third, we develop a generative data engine powered by the \textit{\rapLongName (\rapName)} model, which synthesizes 0.5B diverse training instances through a dual-path cross-verification pipeline, reducing label noise by 80.5\% compared to conventional approaches. Extensive experiments demonstrate that \modelName achieves state-of-the-art performance on open-world detection benchmarks while significantly expanding semantic coverage beyond fixed-vocabulary constraints. Our work establishes a new paradigm for scalable multimodal detection and data generation in open-world scenarios. Data\&Code are available at \url{https://github.com/WeChatCV/WeVisionOne}.

\end{abstract}
\vspace{-1em}

\section{Introduction}
\label{sec:intro}
    Recent advancements in vision models~\cite{ovdistill, pure, unig, wudi, ppt, mllmerg} have focused on unifying diverse prompting strategies—such as text, visual, or hybrid inputs—to enable versatile object understanding and segmentation. Pioneering works like \cite{seem, dinov, trex2, mask-dino} have demonstrated the potential of multimodal prompting for tasks ranging from open-set detection to segmentation. Despite their emerging capabilities, open-world detection and segmentation methods leveraging multimodal prompts demonstrate inherent technical deficiencies.1) \textit{Late-stage fusion of multimodal features}. \cite{seem, dinov} process text, visual or other prompts, and backbone features in separate subspaces, merging them only in later encoding stages. \cite{trex2} similarly adopts contrastive learning to align modalities without joint optimization at the encoding stage. This delayed interaction restricts cross-modal alignment, particularly in ambiguous scenarios where early fusion of text and visual cues could disambiguate object semantics. 2) \textit{Insufficient exploration of query selection mechanisms}. Most of these methods employ generic cross-attention for text-guided query selection, or handle queries corresponding to text and visual prompts independently, none of which explicitly optimizes the structural alignment between queries corresponding to text and visual cues respectively. Like \cite{trex2} processes text and visual prompts independently. This results in the selection of sub-optimal queries that do not promote optimization between text and images.

    Beyond these architectural limitations, a broader challenge persists across the field: the reliance on caption-derived vocabularies or fixed-vocabulary, which inherently constrain generalization to novel or composite objects.~\cite{seem, dinov, trex2} derive their object vocabularies from entity words extracted from image captions or use fixed-vocabulary detector(like TAP~\cite{tap}).  While this approach avoids manually predefined categories, it remains bounded by the entities explicitly mentioned in the training captions. For instance, TRex-2~\cite{trex2}  employ TAP~\cite{tap} to annotate SA-1B~\cite{sam} with a category name from a dictionary of 2560 classes. Nevertheless, the reality that SA-1B~\cite{sam} is associated with a vocabulary significantly exceeding 2560 substantially constrains the dataset's potential.

    In this work, we propose \modelName, a novel text-guided visual Prompt DINO, that addresses these challenges through two architectural innovations and a paradigm-shifting data engine. We pioneer an  \textit{early fusion mechanism} that deeply integrates text, visual, and backbone features at the initial encoding stage. Unlike~\cite{seem, dinov, trex2}, our model unifies multimodal signals into a cohesive representation from the outset. This enables richer cross-modal interactions, reducing ambiguity in open-set scenarios. And we introduce  \textit{order-aligned query selection}, a novel paradigm for DETR-based architectures that explicitly optimizes the alignment between text queries and visual queries during decoding. By enhancing semantic correlation and spatial consistency within query priorities, our method facilitates the optimization of text and image query selection in a unified direction, thereby surpassing the heuristic cross-attention mechanisms employed in prior research. Also, we overcome the fixed vocabulary bottleneck through a generative data engine with \textit{dual-path cross-verification pipeline}. We proposal an novel \textit{\rapLongName (\rapName)} model, a 0.5B-parameter multimodal large language model(MLLM) that generates semantic labels for input images and their corresponding masks. Our engine leverages \rapName to synthesize 0.5B diverse training instances with enhanced semantic relevance. To ensure label fidelity, we further implement a novel dual-path cross-verification pipeline that substantially reduces label noise by 80.5\% compared to conventional single-path approaches. This integrated data engine pioneers scalable open-world dataset generation, bridging limited annotation vocabularies with real-world detection needs.

    To summarize, our contributions are threefold:
    \begin{itemize}
        \item We propose  an  early fusion mechanism that integrates text, visual, and backbone features during initial encoding stage, unifying multimodal signals into a cohesive representation. 
        \item We introduce order-aligned query selection, a new paradigm for DETR-based architectures that explicitly optimizes the alignment between text queries and visual queries during decoding.
        \item We develop a generative data engine powered by our \rapName, incorporating a dual-path cross-verification pipeline. This approach synthesizes diverse and semantically relevant training data, and establishes a new paradigm for scalable open-world data generation.
    \end{itemize}

\vspace{-0.5em}
\section{Related Work}
\begin{figure*}[h]
    \centering
    \includegraphics[width=1\linewidth]{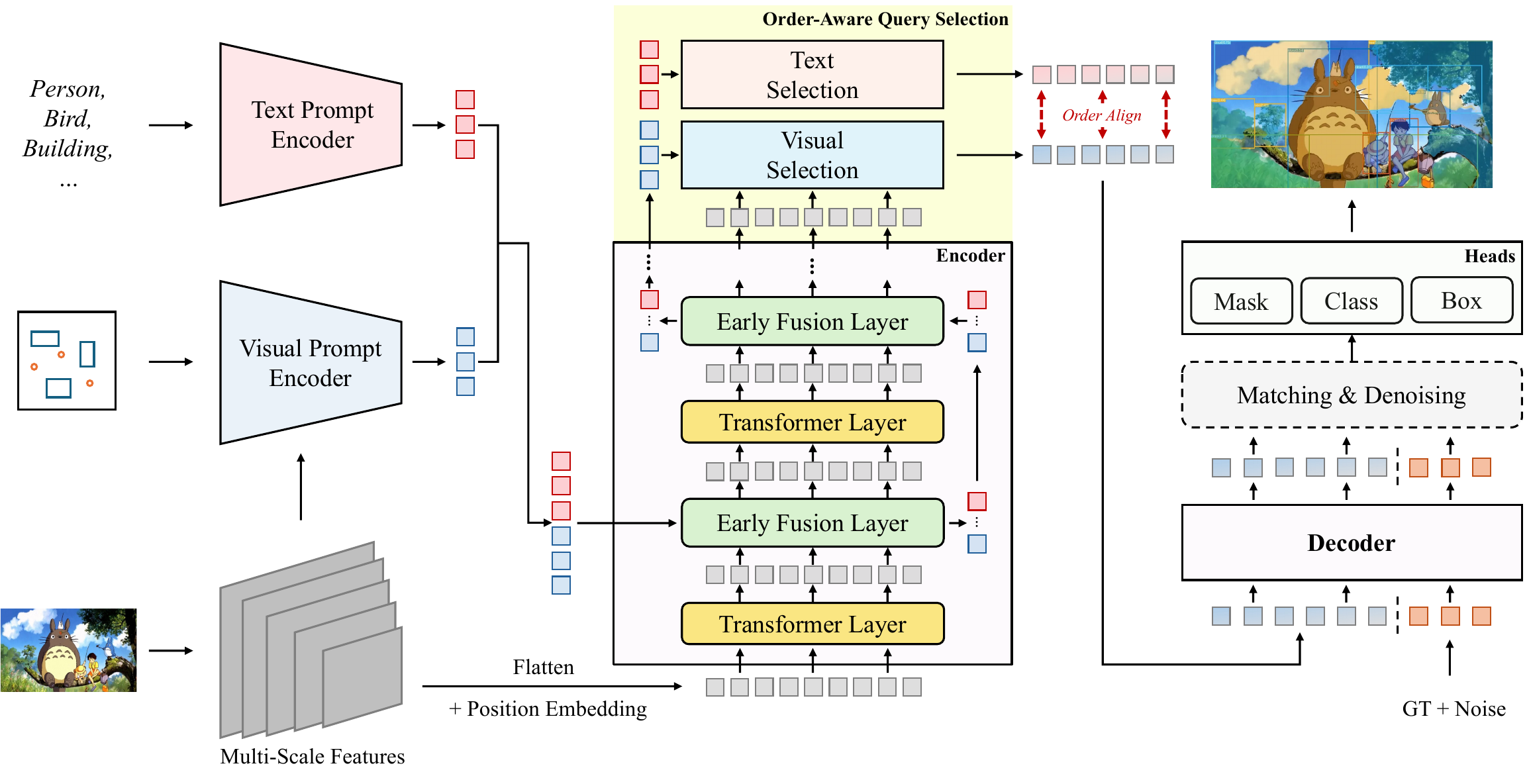}
    \vspace{-2em}
    \caption{\textbf{The Overall Framework of \modelName.} \modelName is constructed based on the DINO architecture. Images are processed by ViT to extract multi-scale features, while visual and text prompts are processed by deformable attention transformer layers and CLIP's text encoder, respectively, to derive embeddings. Through early fusion layers in the backbone encoder and perception-aware query selection, \modelName perform deep fusion and alignment of visual and text in two steps, with the final predictions generated by the Decoder.}
    \label{fig:pipeline}
    \vspace{-1.5em}
\end{figure*}
\vspace{-0.5em}
\subsection{General Recognition via Prompts}
    Recent advances in general recognition models (e.g., detection/segmentation~\cite{ghiasi2022scaling, gu2021open, kamath2021mdetr, minderer2022simple, zhou2022detecting, yao2022detclip, glip, glipv2, grounding-dino}) have demonstrated that multimodal alignment between vision and language modalities serves as the cornerstone for open-world understanding. These frameworks typically employ joint text-visual prompting mechanisms to achieve object localization through aligned cross-modal feature spaces.
     
    \textbf{Text Prompt Paradigm.} Existing methods predominantly focus on text-based cross-modal alignment. GLIP \citep{glip} pioneers the adaptation of image-text pretraining to downstream detection tasks through deep feature fusion, establishing language-aware visual representations. GroundingDINO \citep{grounding-dino} further extends this paradigm by demonstrating that multi-stage cross-attention fusion significantly enhances performance in open-vocabulary detection. While these approaches achieve remarkable progress, their heavy reliance on textual semantics limits their capacity to handle visual concepts with ambiguous linguistic descriptions.
    
    \textbf{Visual Prompt Paradigm.} Complementary to text prompts, visual prompting (points/boxes/masks) provides geometric guidance for specific object localization. SAM \citep{sam} establishes a milestone in interactive segmentation through decoupled visual prompt encoding, though lacking semantic awareness. Subsequent works like SEEM \citep{seem} and DINOv \citep{dinov} enhance this paradigm by integrating semantic context and exemplar-based reasoning. A notable hybrid approach is T-Rex2 \citep{trex2}, which combines both modalities but still treats visual prompts as auxiliary geometric cues. Our key insight lies in that existing methods inadequately exploit the semantic synergy between visual prompts and language guidance, leaving visual prompting's full potential underdeveloped for language-aware feature learning.

    \subsection{Data Engine for Open-World Learning}
    Scalable data curation has become pivotal for training robust open-world detector/segmentation models. SAM2 \citep{sam2} introduces a cyclic model-annotator collaboration framework, iteratively refining the SA-V dataset through human-machine interactions. Alternative strategies employ cross-model verification: YOLO-World \citep{yolo-world} curates CC3M-Lite by relabeling CC3M \citep{cc3m} with GLIP and CLIP, while T-Rex2 \citep{trex2} propagates pseudo-labels across SA-1B and LAION400M \citep{clip}.
    Current methods face three key limitations: (1) Over-reliance on noun phrase extraction from captions misses more detailed concepts and introduces too many redundant nouns; (2) Single-path verification propagates annotation errors; (3) Semantic ambiguity in free-form text introduces label noise. Our data engine addresses these issues and enables the automatic generation of high-quality training data with reduced noise.
\vspace{-1em}
\section{Method}
    Motivated by the limitations observed in existing hybrid prompting methods for detection and segmentation tasks, we introduce \modelName, an end-to-end generic segmentation model built upon the DINO architecture~\cite{dino}. Our model natively integrates multimodal prompting capabilities for both textual and visual inputs while maintaining architectural simplicity. In this section, we first formalize the promptable segmentation paradigm and outline the core design prompt encoders of \modelName(\S\ref{subsec:promptable_task}). We then systematically address three critical challenges: (1) early fusion with visual prompts through cross-modal feature alignment(\S\ref{subsec:early_fusion}), (2) order-aware query selection for maintaining semantic consistency(\S\ref{subsec:query_select}) and (3) construction of a scalable data engine for open-world generalization(\S\ref{subsec:data_engine}). Finally, we introduce the training objective(\S\ref{subsec:train_obj}).
    
    \subsection{Promptable Segmentation Task}
    \label{subsec:promptable_task}
    The \modelName processes an input image $\mathcal I$, with accompanying visual prompts $\mathcal P^v=\{ \mathcal P^v_1,...,\mathcal P^v_K \}$, and textual prompts $\mathcal P^t=\{ \mathcal P^t_1,...,\mathcal P^t_K \}$, where $K$ is the number of interactive prompts per image. Following established architectures in DETR \citep{DETR} and DINO \citep{dino}, our system first extracts hierarchical visual features $F=\{f_1,...,f_L \}$ through a vision backbone network (e.g., ViT~\cite{vit} or Swin Transformer~\cite{swin}), where $L$ represents the number of feature pyramid levels.
    
    \textbf{Visual Prompt Encoder.} Specifically, our visual encoding module employs $L$ transformer blocks with deformable attention mechanisms \citep{dinov}. For each target object, we initialize a learnable query embedding $q_i$ and progressively refine it through multi-scale feature aggregation. The transformation at each layer $l$ can be formulated as:
    \begin{equation}
    \label{eq1}
        \smash{v^l_i = \textit{MSDA}(v^{l-1}_i, f_l), l \in [1,...,L],\quad\text{with}\quad v^0_i=q_i} \\
    \end{equation}
    where \(\textit{MSDA}(\cdot)\) denotes multi-scale deformable attention operation. The final visual prompt representations $V=\{ v_1,...,v_K \}$ are obtained as $v_i=v^L_i$.
    
    \textbf{Text Prompt Encoder.} Building upon insights from recent open-world detection studies, we employ a CLIP~\cite{clip} text encoder rather than conventional BERT~\citep{bert} approaches used in earlier models like ~\cite{grounding-dino,glip}. This design choice enables better remantic correlation modeling for input words, phrases, and short sentences, generating text prompts $T=\{ t_1,...,t_K \}$ that effectively guide cross-modal feature alignment. 

    The integrated representation (text($T$), visual($V$), multi-scale features($F$)) feeds into the backbone encoder for multimodal fusion, enabling dynamic adaptation to prompts while preserving cross-modal alignment.
    
    \subsection{Early Fusion with Visual Prompts}
    \label{subsec:early_fusion}
\begin{figure}[t!]
        \centering
        \includegraphics[width=1\linewidth]{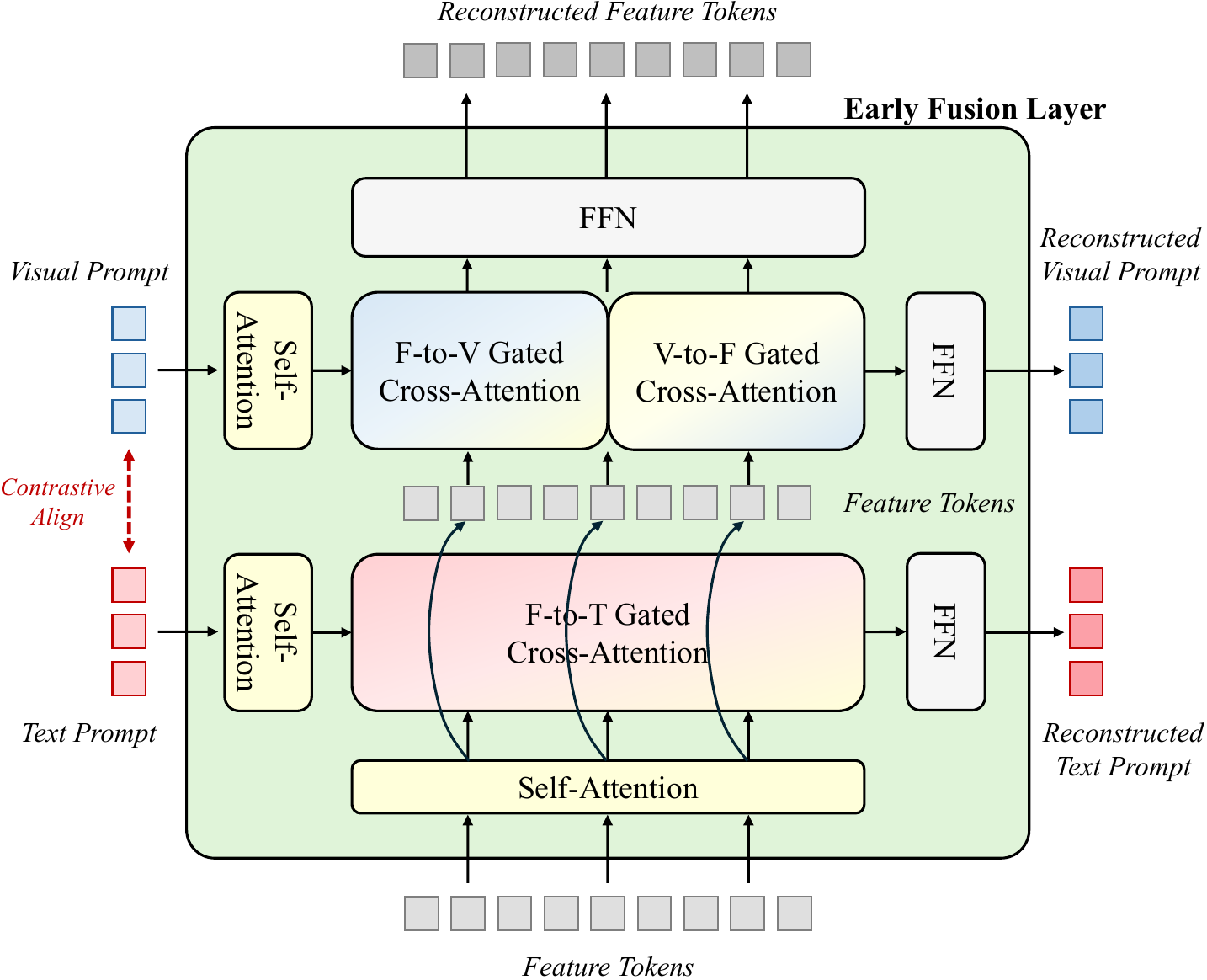}
        \vspace{-2em}
        \caption{\textbf{Structure of Early Fusion Layer.} Notably, the contrastive alignment of visual and text prompts is completed before early fusion. By employing Gated Cross-Attention, text prompts are reconstructed with image features, while visual prompts and image features mutually reconstruct each other.}
        \label{fig:early_fusion}
        \vspace{-1.5em}
    \end{figure}

    We are the first to propose the early fusion framework that integrates text prompts, visual prompts, and input image features through multi-modal interaction. While this fusion mechanism effectively reduces the modality gap between textual and visual features, enhances prompt-image correlation, and improves target recall, We found unintended consequences of severe hallucinations in our experiments.  This phenomenon stems from the inherent limitation of conventional cross-attention mechanisms: they compulsively reconstruct Key features most similar to the Query features, even when all Key features are substantially dissimilar to the current Query. To address this non-target feature reconstruction issue, we introduce a novel gated cross-attention mechanism incorporating a learnable background token ($B$). Reconstruction occurs exclusively when the Query-Key similarity exceeds an adaptive threshold; otherwise, the background token automatically activates to suppress hallucination. As illustrated in Fig.~\ref{fig:early_fusion}, our architecture comprises three core components: 1) a self-attention layer for intra-modal refinement, 2) the proposed gated cross-attention layer for inter-modal interaction, and 3) a feedforward neural network (FFN) for feature transformation. The cross-attention module implements three interaction pathways: feature-to-visual prompt, visual prompt-to-feature, and feature-to-text prompt. For visualization clarity, Fig.~\ref{fig:early_fusion} presents only stage-2 features mapped to text space through our two-phase training scheme (detailed in \S\ref{subsec:train_obj}). The enhanced cross-modal fusion operates through the following formulation:
    \begin{equation}
    \label{eq2}
    \begin{split}
        \smash{t_i' = \mathrm{GatedAttn}(f_i,t_i,t_i)} \\
        \smash{v_i' = \mathrm{GatedAttn}(f_i,v_i,v_i)} \\
        \smash{f_i' = \mathrm{GatedAttn}(v_i,f_i,f_i)}
    \end{split}
    \end{equation}
    where the gated attention mechanism is defined as:
    \begin{equation}
    \label{eq2}
        \smash{\mathrm{GatedAttn}(Q,K,V) = \mathrm{softmax}(\textstyle\frac{Q \cdot [K, B]}{\sqrt{d_k}})[V,B]}
    \end{equation}
    
    Here, $B$ denotes our learnable background token. Following TRex-2~\cite{trex2} methodology, we employ region-level contrastive learning to align visual and textual modalities, utilizing symmetric cross-entropy loss~\cite{clip} for embedding alignment:
    \vspace{-0.8em}
    \begin{equation}
    \label{eq3}
        \smash{\mathcal L_{align} = - \frac{1}{K} \sum_{i=1}^{K}\mathrm{log} \frac{v_i \cdot t_i}{\sum_{j=1}^{K} v_i \cdot t_j}}
    \end{equation}
    In our experience, symmetric cross-entropy losses have better convergence than InfoNCE loss~\cite{infonceloss}.
    
    In the context of training detection and segmentation tasks, it is common practice to construct negative samples. For negative prompt construction, while text prompts naturally leverage category vocabularies for negative sampling, visual prompts require explicit negative construction. We address this by aggregating visual prompt features across all categories within each batch and computing inter-category averages to generate negative visual prompts. This approach ensures sufficient diversity in negative samples while maintaining semantic relevance.

    To mitigate annotation granularity discrepancies across datasets (e.g., "bird" in Dataset A vs. "pigeon" in Dataset B), we implement a dataset-aware sampling strategy. Our novel data sampler ensures each training batch contains only intra-dataset samples, preventing semantically similar but categorically distinct concepts from becoming conflicting negative samples. This is particularly critical in visual-prompt open-world detection/segmentation scenarios where label hierarchies may differ substantially across source datasets.

    \subsection{Order-aware query selection}
    \label{subsec:query_select}

    \modelName adopts a DETR-based architecture that selects top-K encoder features through similarity computation between encoder features and prompt embeddings, following DINO~\cite{dino} and MaskDINO~\cite{mask-dino}. However, our approach fundamentally diverges from existing methods by simultaneously considering both textual-visual query similarity and inter-query correlations with visual prompts. To further strengthen the alignment between visual and textual prompts, we introduce order-aware alignment during query selection, which establishes the second-level alignment between visual and textual prompts during training. Specifically, we employ Kendall’s $\tau$ coefficient to quantify the ordinal consistency~\cite{rankloss} between queries selected by visual prompts and text prompts, the Kendall’s $\tau$ coefficient can be expressed as:
    \vspace{-1em}
    \begin{equation}
    \label{eq5}
        \smash{\tau = \frac{P_{c}-P_{d}}{\frac{1}{2}N(N-1)}}
    \end{equation}
    where $P_{c}$ represents concordant pairs and $P_{d}$ represents discordant pairs, $N$ represents the total number of queries. Given the similarity score between query embeddings and prompt embeddings $S^v={s^v_1,...,s^v_N}$ and $S^t={s^t_1,...,s^t_N}$, we calculate $P_{c}$ and $P_{d}$ as follows:
    \begin{equation}
    \label{eq6}
        \smash{\tau = \frac{\sum_{i}\sum_{j<i} \mathrm{sgn}(s^{t}_{i}-s^{t}_{j}) \cdot \mathrm{sgn}(s^{v}_{i}-s^{v}_{j})}{\frac{1}{2}N(N-1)}}
    \end{equation}
    Since the $\mathrm{sgn}$ is not differentiable, we approximate the sign function with the $tanh$ function and utilize the opposite of $\tau$ as the order alignment loss: 
    \begin{equation}
    \label{eq7}
        \smash{\mathcal L_{order} = -\frac{\sum_{i}\sum_{j<i} \mathrm{tanh}(s^{t}_{i}-s^{t}_{j}) \cdot \mathrm{tanh}(s^{v}_{i}-s^{v}_{j})}{\frac{1}{2}N(N-1)}}
    \end{equation}

    Our proposed alignment mechanism operates at two complementary levels: The primary alignment directly optimizes the learning objective through $\mathcal{L}_{align}$ (as given in Eq.~\ref{eq3}), while our novel order-aware alignment provides indirect assurance through two critical aspects: (1) Semantic proximity between corresponding text and visual prompts for candidate queries naturally reduces overall training complexity by minimizing feature space discrepancies; (2) Training-inference consistency preservation ensures equivalent substitutability between text and visual prompts during inference through maintained representational parity.

    Fig.~\ref{fig:order-analysis} demonstrates the effectiveness of our proposed method. Prior to applying the order-aware loss optimization, we observe significant disorder in both textual and visual prompt queries, with no discernible alignment between corresponding semantic representations. The optimization process induces a remarkable transformation - post-optimization analysis reveals substantially enhanced consistency between semantically equivalent textual and visual prompt queries. 

    \subsection{Data Engine}
    \label{subsec:data_engine}
    Effective visual concept learning from rich image-text pairs is fundamental to enhancing the generalization capabilities of open-vocabulary object detectors. Existing methods predominantly rely on captioning models(\cite{blip, llava, sharegpt4v} etc.), coupled with lexical analysis tools like SpaCy~\cite{spacy} to extract noun phrases. These noun phrases are subsequently employed to generate pseudo-labels via open-set detection models(~\cite{grounding-dino, glip}). Nonetheless, these techniques have significant drawbacks. They frequently omit valid labels, introduce erroneous annotations, and are susceptible to irrelevant lexical noise. All of these issues detrimentally affect detection performance.

    To address these challenges, we propose a novel \textit{\rapLongName (\rapName)} model, a 0.5B MLLM, which significantly enhances annotation accuracy. Our framework further integrates an innovative dual-path cross-validation pipeline that substantially reduces label noise. This comprehensive approach enables the fully automated generation of high-quality open-world detection datasets while preserving the fidelity of annotations.

    \begin{figure*}[t!]
        \centering
        \includegraphics[width=1\linewidth]{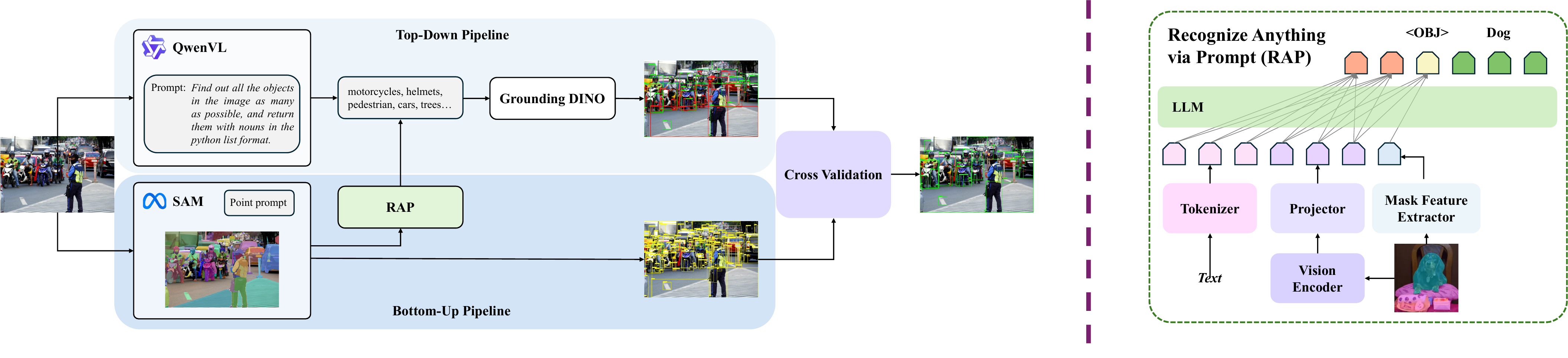}
        \vspace{-2em}
        \caption{\textbf{Data Engine and RAP Frameworks.} \textbf{Left:} The data engine with dual-path cross-verification pipeline. \textbf{Right:} The proposed Recognize Anything via Prompting (RAP), a 0.5B-parameter MLLM model.}
        \label{fig:data-engine}
        \vspace{-1.5em} 
    \end{figure*}

    \textbf{\rapLongName.} Despite the remarkable capabilities of the Segment Anything Model (SAM)~\cite{sam} in detecting a wide array of objects within images, and the substantial progress achieved in region-level understanding through multimodal models~\cite{Kosmos-2, rasheed2024glamm, yuan2024osprey, chen2023position, chen2023shikra}, a critical question persists: Can we effectively assign semantic labels to arbitrary mask prompts? To address this challenge, we introduce \rapName, a novel framework that bridges visual segmentation with semantic recognition. As illustrated in Fig. \ref{fig:data-engine}, our \rapName model processes raw image inputs alongside their corresponding object masks to generate precise semantic annotations for the specified regions. The proposed \rapName comprises three key components: 
    
    1. Feature extraction. We perform masked pooling over multi-layer image features to extract region-specific features corresponding to input masks, and concatenate the mask token embeddings obtained via a projection layer with the original image tokens to form a unified representation. This preserves both global contextual information from the complete image and localized object-specific features within masked regions, effectively integrating scene-level understanding with instance-level visual cues.
    
    2.Ground Data Cleaning. To standardize the format of our ground truth data, we leveraged the Qwen model~\cite{qwen2} to eliminate redundant descriptive phrases through comprehensive text normalization. Specifically, we transformed verbose expressions into their canonical semantic equivalents (e.g., converting ``any part of the sky" to ``sky"). This preprocessing strategy serves dual purposes: (1) it reduces the \rapName's learning complexity by minimizing lexical variability, and (2) it mitigates the risk of hallucinations by ensuring precise semantic alignment between textual annotations and visual content.

    3.Training. The proposed methodology employs a two-stage training strategy. During the first stage, we conduct pre-training using 10M samples, specifically optimizing only the projector while keeping other model components frozen. The second stage implements instruction-tuning, where we unfreeze and jointly optimize the entire model to enhance task-specific adaptation capabilities.
    
    \textbf{Dual-path cross-verification pipeline.}We propose a dual-pipeline annotation framework combining bottom-up and top-down approaches with cross-verification filtering, consisting of three key components:
    
    1. Top-Down Pipeline. Our top-down pipeline leverages the Qwen-VL model~\cite{Qwen2-VL} to extract semantic tags from images. These tags are subsequently fed into an open-set detection model to generate corresponding bounding boxes. This pipeline demonstrates strong capability in detecting large-scale salient objects (e.g., trees, roads, sky).
    
    2. Bottom-Up Pipeline. To address the inherent limitation of top-down pipeline in capturing fine-grained details, we implement a complementary bottom-up strategy using the \rapName integrated with SAM~\cite{sam}. This combination enables effective detection of smaller objects and detailed components (e.g., eyes, noses, windows) that are frequently missed by conventional top-down pipeline.
    
    3. Cross-Verification. The detection outputs from both pipelines undergo rigorous validation through a three-stage process. First, We employ the Hungarian matching algorithm to establish correspondences between bounding boxes from different pipelines. Next, we utilize the BGE model~\cite{bge_embedding} to compute similarity scores between the tags associated with the matched bounding boxes. Finally, we filter out the candidate pairs whose similarity scores were below the threshold of empirical determination to further clean up the annotated results.

    \subsection{Training Objective}
    \label{subsec:train_obj}
    Building upon previous DINO-like models~\cite{dino}, we employ a combination of L1 loss and GIoU loss~\cite{giou} for bounding box regression, while for mask segmentation, we adopt cross-entropy loss and dice loss~\cite{dice}. Following the methodology established in GroudingDINO~\cite{grounding-dino}, we utilize contrastive loss between predicted objects and text prompts for classification tasks. To accelerate model convergence, we incorporate denoising training as implemented in DINO~\cite{dino}.
    The complete objective function for Prompt DINO can be expressed as:
    \begin{equation}
    \label{eq8}
        \smash{\mathcal{L} = \textstyle \mathcal{L}_{cls} \!+\! \mathcal{L}_{bbox} \!+\! \mathcal{L}_{mask} \!+\! \mathcal{L}_{DN} \!+\! \mathcal{L}_{align} \!+\! \mathcal{L}_{order}}
    \end{equation}
    Our training methodology adopts a progressive multi-stage strategy: we first optimize the text prompt while keeping the visual components fixed, followed by joint optimization of both textual and visual prompt representations.

\section{Experiments}
    \begin{table*}[htbp]
\centering
\footnotesize
\renewcommand\arraystretch{0.9} 
  \centering
  \vspace{-2em}
    \resizebox{\linewidth}{!}{
    
    \begin{tabular}{l|c|c|cc|cc|ccc}
    \toprule
    \multicolumn{1}{c|}{\multirow{2}{*}{Method}} 
    & \multirow{2}{*}{Backbone} 
    & \multirow{2}{*}{Prompt Type} 
    & \multicolumn{2}{c|}{COCO (in-domain)} 
    & \multicolumn{2}{c|}{LVIS (in-domain)} 
    & \multicolumn{3}{c}{ADE20K (out-domain)} \\
    & & & mask AP & box AP & mask AP & box AP & PQ & mask AP & box AP \\
    \midrule
    Mask2Former \citep{mask2former} & Swin-L & Close-Set 
    & 48.6 & 52.1 & - & - & - & - & - \\
    OneFormer \citep{oneformer} & Swin-L & Close-Set 
    & 48.9 & - & - & - & - & - & - \\
    MaskDINO\textsuperscript{*} \citep{mask-dino} & Swin-L & Close-Set 
    & 50.6 & 56.2 & - & - & - & - & - \\
    kMaX-DeepLab \citep{kmax} & Swin-L & Close-Set 
    & 58.1 & - & - & - & - & - & - \\
    \midrule
    GLIPv2 \citep{glipv2} & Swin-H & Text Prompt 
    & 48.9 & - & - & - & - & - & - \\
    X-Decoder \citep{xdecoder} & DaViT-L & Text Prompt 
    & 46.7 & - & - & - & 21.8 & 13.1 & 38.1 \\
    OpenSeeD \citep{openseed} & Swin-L & Text Prompt 
    & 53.2 & 58.2 & 21.0 & 23.0 & 19.7 & 15.0 & 17.7 \\
    T-Rex2 \citep{trex2} & Swin-L & Text Prompt 
    & - & 52.2\textsuperscript{*} & - & 45.8\textsuperscript{*} & - & - & - \\
    APE \citep{ape} & ViT-L & Text Prompt 
    & 49.3 & 58.3 & 53.0 & 59.6 & 27.2 & 24.4 & 29.6 \\
    \midrule
    DINOv \citep{dinov} & Swin-L & Visual Prompt 
    & 50.4 & 54.2 & - & - & 23.2 & 15.1 & 14.3 \\
    T-Rex2 \citep{trex2} & Swin-L & Visual Prompt 
    & - & 46.5\textsuperscript{*} & - & 45.3\textsuperscript{*} & - & - & -  \\
    \midrule
    Prompt-DINO & ViT-L & Text Prompt & 51.1 & 58.6 & 56.5 & 60.1 & 29.6 & 24.6 & 28.6 \\
    Prompt-DINO & ViT-L & Visual Prompt & 51.0 & 58.0 & 55.6 & 59.9 & 35.9 & 28.0 & 32.7 \\
    \bottomrule
    \end{tabular}
    
    }
  \vspace{-1em}
  \caption{Generic Segmentation results of \modelName on multiple datasets. "-" denotes the model does not have performance reported or does not have ability for the specific task. "*" denotes the performance reported under different domain settings. MaskDINO uses the results reported in DINOv \cite{dinov}}
  \label{tab:generic-seg}
  \vspace{-2em}
\end{table*}
    \subsection{Settings}
    In \modelName, we employ ViT-L~\cite{vit} as the backbone and fine-tune CLIP-L~\cite{clip} as our text prompt encoder. The visual prompt encoder consists of 6 layers of deformable cross-attention transformer layers, with a hidden dimension of 256. We adopt the "two-stage" structure proposed by DINO~\cite{dino}, selecting 900 queries for decoding after the backbone encoder.
    
    \textbf{Datasets.} During the training process, we utilized three categories of datasets: \textit{open-source detection datasets.} These include COCO \cite{coco} with 118K images and LVIS \cite{lvis} with 160K images, both of which have mask annotations. For datasets without mask annotations, such as Object365 \cite{objects365}, OID \cite{oid}, and V3Det \cite{v3det}, we first apply SAM to generate mask annotations. Due to the presence of low-quality data and long-tail distribution in Object365 and OID, which hinder general object recognition, we resampled them, selecting 700K and 1M data points, respectively. \textit{grounding datasets.} This includes the GoldG \cite{goldg} dataset with 100K images. \textit{self-built datasets.} We re-annotate and filter CC3M \cite{cc3m} and SA-1B \cite{sam} using the RAP and dual-path cross-validation data engine introduced in \S \ref{subsec:data_engine}, yielding high-quality datasets of 3M and 7M images, respectively, which constitute our primary training data. For testing, we utilize COCO and LVIS for in-domain evaluation and ADE20K \cite{ade20k}, SeginW \cite{xdecoder}, and Cityscapes \cite{cityscapes} for out-domain evaluation.

    \textbf{Metrics.} We employ Average Precision (AP) as the evaluation metric, utilizing mask AP for instance segmentation, and box AP for object detection. For ADE20K and CtiyScapes, we employ Panoptic Quality(PQ) and mean intersection over union (mIoU) for panoptic segmentation as well. For SegInW, we adopt AP-Average and AP-Median as the evaluation metric.

    \subsection{Generic Segmentation}
    To evaluate the effectiveness and generalization capability of our \modelName, we conduct evaluations on both in-domain and out-domain datasets, with the results presented in Tab. \ref{tab:generic-seg}.

    \textbf{In-Domain Segmentation.} We conduct in-domain experiments on COCO and LVIS. Compared to methods based on visual prompts, our \modelName achieved significantly better results. Compared to the previous work DINOv \cite{dinov}, we improve mask AP by 0.6 and box AP by 3.8 on COCO. Compared to methods based on text prompts, \modelName also demonstrates superior results when using text prompts. For example, compared to APE \cite{ape}, we lead by 3.5 in mask AP and 0.5 in box AP on LVIS.

    \textbf{Out-Domain Segmentation.} We use ADE-20K and Cityscapes as an out-domain dataset and conduct zero-shot testing on it, demonstrating the strong generalization ability of our \modelName. Under the out-domain setting (where data marked with * indicates performance reported under different domain settings), our method achieves the best performance on ADE20K across all methods for both prompt types, whether using visual prompts or text prompts. Compared to the state-of-the-art method APE on ADE20K, we improve PQ by 8.7, mask AP by 3.6, and box AP by 18.4, representing increases of 32.0\%, 14.8\%, and 10.5\%, respectively. Among methods based on visual prompts, compared to the previous method DINOv, we improve PQ by 12.7, mask AP by 12.9, and box AP by 18.4, representing increases of 54.7\%, 85.4\%, and 128.7\%, respectively. Compared to APE on Cityscapes, we also improve PQ by 1.0 and mIoU by 8.0.

    \textbf{Segmentation in the Wild.} To further evaluate the generalization capabilities of our model, we perform zero-shot testing on the SeginW benchmark, which includes 25 datasets. Compared to the previous work DINOv, when using visual prompts, we improve the average AP by 16.2 and the median AP by 14.5. Among text-prompt-based methods, we also surpass APE by 6.8 in average AP and 6.9 in median AP. This further validates the strong generalization ability of our \modelName.
    
    \begin{table}[htbp]
\centering
\footnotesize
  \centering
  \vspace{-2em}
    \resizebox{\linewidth}{!}{
    
    \begin{tabular}{l|c|cc|cc}
    \toprule
    \multicolumn{1}{c|}{\multirow{2}{*}{Method}} 
    & \multirow{2}{*}{Type} 
    & \multicolumn{2}{c}{CityScape}
    & \multicolumn{2}{c}{SegInW} \\
    & & PQ & mIoU & AP-avg & AP-med \\
    \midrule
    kMaX-DeepLab-L \citep{kmax}  & in-domain & 68.4 & 83.5 & - & - \\
    \midrule
    X-Decoder-L \citep{xdecoder}  & out-domain & 38.1 & 52.0 & 22.3 & 32.3 \\
    OpenSeeD-L \citep{openseed}  & out-domain & 41.4 & 47.8 & 36.1 & 38.7\\
    APE-L \citep{ape}  & out-domain & 33.3 & 44.2 & 49.6 & 52.2\\
    DINOv \citep{dinov} & out-domain & - & - & 40.6 & 44.6\\
    \midrule
    Prompt-DINO \citep{ape}  & out-domain & 34.3 & 52.2 & 56.4 & 59.1\\
    \bottomrule
    \end{tabular}
    
    }
  \vspace{-1em}
  \caption{Zero-shot Segmentation results of \modelName on the CityScape \cite{cityscapes} and SegInW \cite{xdecoder} dataset.}
  \label{tab:cityscape}
\end{table}

    \subsection{Effectiveness Analysis}
    \begin{table}[htbp]
\centering
\footnotesize
  \centering
  \vspace{-1em}
    \resizebox{\linewidth}{!}{
    
    \begin{tabular}{cccc|ccc}
    \toprule
    \multicolumn{4}{c|}{Method} 
    & \multicolumn{3}{c}{ADE20K (out-domain)} \\
    Baseline &  Text Align & Early Fusion & Order Align & PQ & mAP & bAP \\
    \midrule
    \checkmark & $\times$ & $\times$ & $\times$ & 25.0 & 20.4 & 23.5 \\
    \checkmark & \checkmark & $\times$ &  $\times$ & 32.7 & 25.5 & 30.5 \\
    \checkmark & \checkmark & \checkmark & $\times$ & 34.7 & 26.7 & 31.7 \\
    \checkmark & \checkmark & \checkmark & \checkmark & 35.9 & 28.0 & 32.7 \\
    \bottomrule
    \end{tabular}
    
    }
  \vspace{-1em}
  \caption{\textbf{Ablation} of proposed components.}
  \label{tab:ablation}
  \vspace{-2em}
\end{table}
    To validate the effectiveness of our proposed components, we conduct ablation studies on Text Prompt Alignment, Early Fusion, Order Alignment, and Data Engine. 
    As our baseline, we implement the naive MaskDINO method with the FC head replaced with a user-prompted visual embedding.
    The ablation results are presented in Tab. \ref{tab:ablation}. 
    
    \textbf{Effectiveness of Early Fusion.} By comparing rows 1 and 2 in Tab. \ref{tab:ablation}, it can be observed that incorporating text align in the backbone encoder (stage one of DINO) increases the mask AP and box AP on ADE20K by 5.1 and 7.0, respectively, and PQ by 7.7, compared to the baseline. This indicates that naive alignment of text and visual can enhance the model's general recognition capability. Comparing rows 2 and 3 in Tab. \ref{tab:ablation} shows that early fusion of visual prompts, guided by text alignment, significantly improves the model's performance, with mask AP and box AP on ADE20K increasing by 1.2 and 1.2, respectively, and PQ by 2.0. This demonstrates that visual prompts effectively complement features that are difficult to describe with text prompts and, under the guidance of text prompts, achieve more generalizable expressive power.

    \begin{figure}[h!]
        \vspace{-0.7em}
        \centering
        \includegraphics[width=1\linewidth]{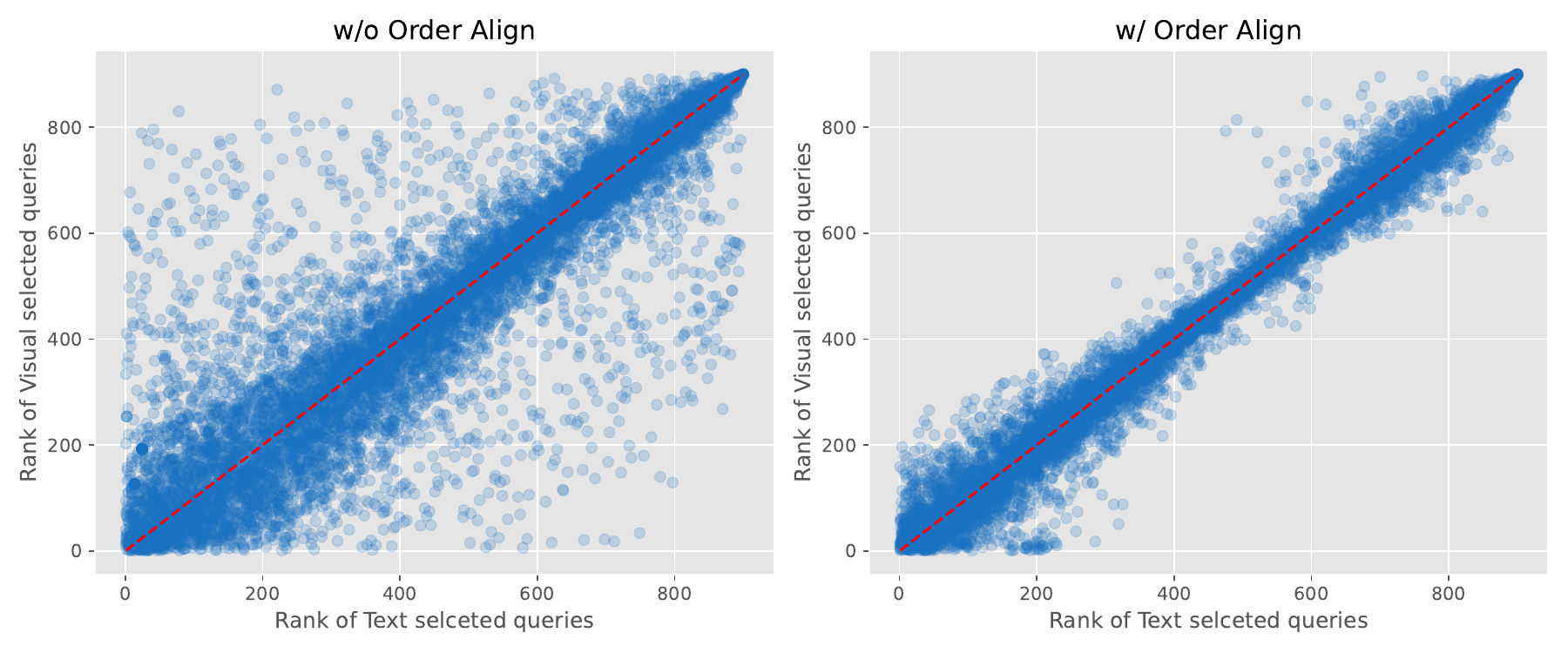}
        \vspace{-2em}
        \caption{Order Alignment Comparison. \textbf{Left:} Visual and text should demonstrate strong correlation, but the queries selected by the two prompts exhibit significant disorder. \textbf{Right:} Order alignment significantly improves the consistency of visual-text query selection and reinforces the alignment between corresponding semantics.}
        \label{fig:order-analysis}
        \vspace{-1em}
    \end{figure}

    \textbf{Effectiveness of Text-guided Visual Prompts.} Rows 1, 2, and 3 in Tab.~\ref{tab:ablation} show that merely adding text alignment provides limited improvement to the model. However, after early fusion to eliminate semantic ambiguity, the two types of prompts can serve as complementary information, providing stronger recognition capabilities. Next, we introduced order alignment during the query selection stage to further guide visual prompts with text. Rows 3 and 4 in Tab.~\ref{tab:ablation} demonstrate that our two-stage guidance deeply aligns visual and text prompts, leading to further performance improvements. Fig.~\ref{fig:order-analysis} also demonstrates that our order alignment method further enhances the alignment between visual and text, resulting in a more consistent selection order for queries.

    \begin{figure}[t!]
        \centering
        \vspace{-2em}
        \includegraphics[width=1\linewidth]{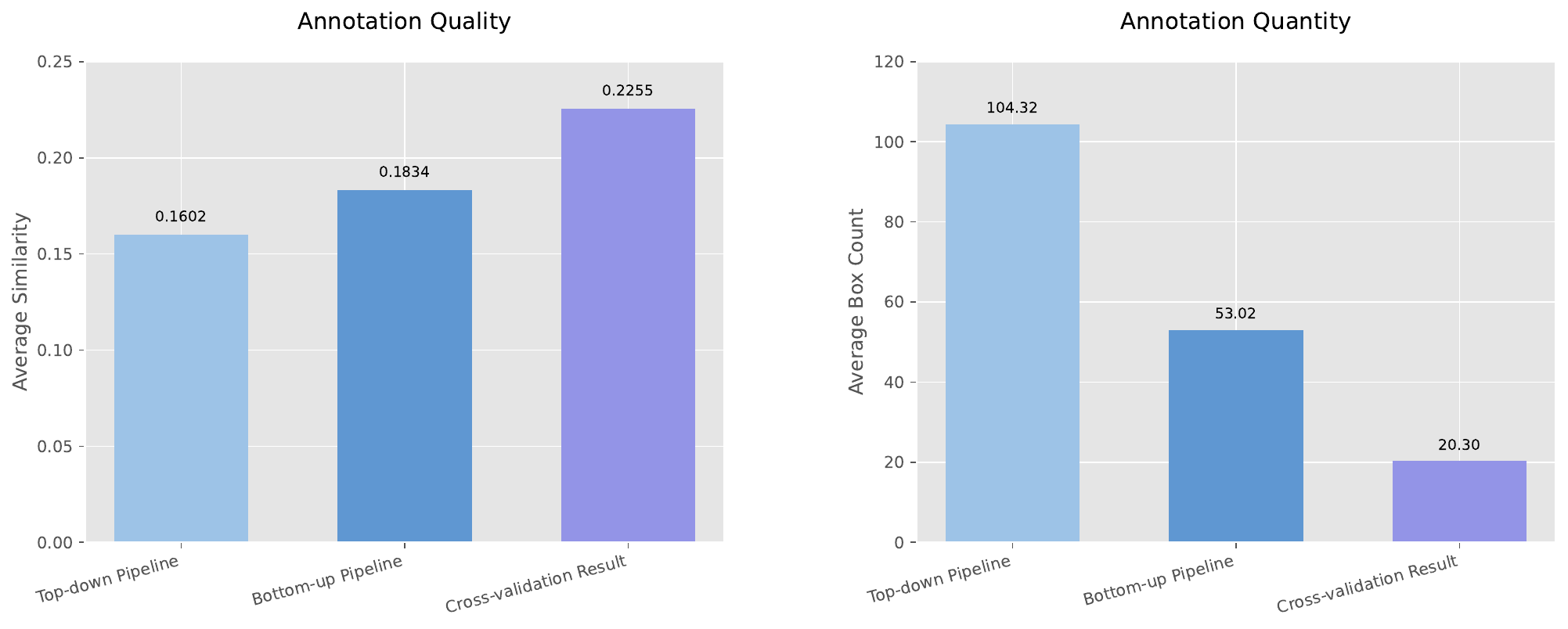}
        \vspace{-2em}
        \caption{Annotation quality and quantity in the data engine.}
        \vspace{-1em}
        \label{fig:anno-analysis}
    \end{figure}
    
    \textbf{Effectiveness of Data Engine.} Within the data engine, we employ a cross-validation module to filter data annotated by the top-down and bottom-up pipelines. As illustrated in Fig.~\ref{fig:anno-analysis}, the cross-validation module filters approximately 80.5\% of instance annotations and enhances the annotation clip similarity from 0.16 to 0.23, highlighting the effectiveness of the cross-validation module in generating high-quality data. As shown in Tab.~\ref{tab:r3}, incorporating data generated by our data engine into the training set improves \modelName's box AP by 12\% on LVIS and 5.9\% on ADE20K. To further demonstrate the effectiveness of RAP, we process a subset of SA-1B using different data engines in Tab.~\ref{tab:r2}, revealing that RAP enhances model performance by 1.6\% mask AP and 2.6\% box AP on ADE20K. This indicates that the proposed data engine effectively enhances data quality, while RAP generates more diverse data, significantly improving the model's generalization capability.
    \begin{table}[htbp]
\centering
\footnotesize
  \centering
  \vspace{-2em}
    \resizebox{\linewidth}{!}{
    
    \begin{tabular}{l|cc|ccc}
    \toprule
    \multicolumn{1}{c|}{\multirow{2}{*}{Method}} 
    & \multicolumn{2}{c|}{LVIS (in-domain)}
    & \multicolumn{3}{c}{ADE20K (out-domain)} \\
    & mask AP & box AP & PQ & mask AP & box AP \\
    \midrule
    w/o DataEngine & 46.0 & 47.9 & 30.8 & 23.4 & 26.8 \\
    w Data Engine & 55.6 & 59.9 & 35.9 & 28.0 & 32.7 \\
    \bottomrule
    \end{tabular}
    
    }
  \vspace{-3mm}
  \caption{\textbf{Ablation} of Data Engine (Visual Prompt Results). }
  \label{tab:r3}
  \vspace{-5mm}
\end{table}
    \begin{table}[htbp]
\centering
\footnotesize
  \centering
  \vspace{-1em}
    \resizebox{\linewidth}{!}{
    
    \begin{tabular}{l|cc|ccc}
    \toprule
    \multicolumn{1}{c|}{\multirow{2}{*}{Dataset}} 
    & \multicolumn{2}{c|}{COCO (in-domain)}
    & \multicolumn{3}{c}{ADE20K (out-domain)} \\
    & mask AP & box AP & PQ & mask AP & box AP \\
    \midrule
    COCO+GoldG & 49.8 & 55.7 & 18.2 & 11.0 & 11.6 \\
    COCO+SA-1B (QwenVL) & 50.1 & 56.7 & 22.3 & 17.8 & 20.4 \\
    COCO+SA-1B (QwenVL + RAP) & 50.8 & 58.1 & 22.5 & 19.4 & 23.0 \\
    \bottomrule
    \end{tabular}
    
    }
  \vspace{-3mm}
  \caption{\textbf{Ablation} of Training Dataset (Text Prompt Results). For efficient training, we use a 2M subset of SA-1B.}
  \label{tab:r2}
  \vspace{-5mm}
\end{table}

\vspace{-1em}
\section{Conclusion}
\vspace{-0.5em}
    We present \modelName, a text-guided visual Prompt DINO for generic segmentation via three innovations: 1) Early fusion of text, visual, and backbone features during initial encoding to resolve cross-modal ambiguity; 2) Order-aligned query selection for semantic-spatial consistency between text and visual cues in DETR decoding; 3) A generative data engine (\rapName + dual-path verification) synthesizing 0.5B high-fidelity training instances, reducing label noise by 80.5\%. \modelName sets new state-of-the-art results while enabling scalable adaptation to real-world object diversity.

\section*{Acknowledgements}
This work was partially supported by the National Key R\&D Program of China (2022YFB4701400/4701402), SSTIC Grant(KJZD20230923115106012, KJZD20230923\break114916032, GJHZ20240218113604008), National Natural Science Foundation of China (No. 62106149).

{
    \small
    \bibliographystyle{ieeenat_fullname}
    \bibliography{main}
}

\end{document}